\pdfoutput=1

\documentclass[11pt]{article}

\usepackage[final]{acl}

\usepackage{times}
\usepackage{latexsym}
\usepackage{multicol}

\usepackage[T1]{fontenc}

\usepackage[utf8]{inputenc}

\usepackage{microtype}
\usepackage{amsmath}
\usepackage{cleveref}
\usepackage{multirow}
\usepackage{array} 
\usepackage{xcolor} 
\usepackage{caption,subcaption}

\usepackage{inconsolata}

\usepackage{tcolorbox}
\newcommand{\Sref}[1]{\S\ref{#1}}
\newcommand{\dashifted}{\raisebox{0.5\depth}{\tiny$\downarrow$}}

\newcommand{\da}[1]{{\scriptsize\hlprimarytab{\dashifted{#1}}}}

\definecolor{c1}{cmyk}{0,0.6175,0.8848,0.1490} 
\definecolor{c2}{cmyk}{0.1127,0.6690,0,0.4431} 
\definecolor{c3}{rgb}{0.55, 0.71, 0.0}
\definecolor{c4}{cmyk}{0.6765,0.2017,0,0.0667} 
\definecolor{c5}{cmyk}{0,0.8765,0.7099,0.3647} 

\newtcbox{\hlprimarytab}{on line, rounded corners, box align=base, colback=red!10,colframe=white,size=fbox,arc=3pt, before upper=\strut, top=-2pt, bottom=-4pt, left=-2pt, right=-2pt, boxrule=0pt}
\newtcbox{\hlsecondarytab}{on line, box align=base, colback=c3!10,colframe=white,size=fbox,arc=3pt, before upper=\strut, top=-2pt, bottom=-4pt, left=-2pt, right=-2pt, boxrule=0pt}
\newtcbox{\hlmedium}{on line, box align=base, colback=gray!10,colframe=white,size=fbox,arc=3pt, before upper=\strut, top=-2pt, bottom=-4pt, left=-2pt, right=-2pt, boxrule=0pt}

\usepackage{graphicx}

\title{Evaluating Differentially Private Synthetic Data Generation in High-Stakes Domains}

\author{Krithika Ramesh \\
  Johns Hopkins University \\
  \texttt{kramesh3@jh.edu} \\\And
  Nupoor Gandhi \\
    Carnegie Mellon University \\
  \texttt{nmgandhi@cs.cmu.edu} \\\And
    Pulkit Madaan \\
      Johns Hopkins University \\
  \texttt{pmadaan2@jhu.edu} \\\AND
  Lisa Bauer \\
  Amazon \\
  \texttt{bauerlb@amazon.com} \\\And
  Charith Peris \\
  Amazon \\
  \texttt{perisc@amazon.com} \\\And
  Anjalie Field  \\
    Johns Hopkins University \\
  \texttt{anjalief@jhu.edu} \\}

\begin{document}
\maketitle

\begin{abstract}
The difficulty of anonymizing text data hinders the development and deployment of NLP in high-stakes domains that involve private data, such as healthcare and social services. Poorly anonymized sensitive data cannot be easily shared with annotators or external researchers, nor can it be used to train public models. In this work, we explore the feasibility of using synthetic data generated from differentially private language models in place of real data to facilitate the development of NLP in these domains without compromising privacy. In contrast to prior work, we generate synthetic data for real high-stakes domains, and we propose and conduct use-inspired evaluations to assess data quality. Our results show that prior simplistic evaluations have failed to highlight utility, privacy, and fairness issues in the synthetic data. Overall, our work underscores the need for further improvements to synthetic data generation for it to be a viable way to enable privacy-preserving data sharing. 
\end{abstract}
\section{Introduction}
The widespread availability of public digitized text has greatly facilitated the advancement of natural language processing (NLP).
Text processing could also be extremely valuable for processing high-stakes private data, like healthcare records \citep{panchbhai2022systematic}, social workers' notes \citep{gandhi-etal-2023-annotating}, or legal documents \cite{zhong-etal-2020-nlp}.
However, the need to maintain data privacy hinders the responsible development and deployment of models in these domains.

Building NLP tools often requires sharing data externally with contractors or researchers, as agencies like child protective services typically do not have in-house AI expertise.
While data sharing has been accomplished through data use agreements with individual teams
 or laboriously redacting identifiable information from text (e.g., \citet{johnson2016mimic2}), these approaches have limitations. 
Limited sharing still requires increasing the number of people who have access to sensitive data, and it  precludes the development of public benchmarks, which have proved crucial for standardizing model development.
Redaction fails to fully prevent re-identification, as even lower dimensional data is often possible to re-identify with just small amounts of auxiliary data \cite{netflixprize, sweeney-simple}. Furthermore, redacted data is not useful for tasks requiring sensitive information, such as developing a model to identify contact information for potential caretakers of a child \citep{field2023examining}. 

In our work, we propose and conduct use-inspired evaluations of the feasibility of using synthetic data to address these limitations. 
Recent work has proposed sharing synthetic text generated from differentially private language models in place of real data \cite{yue-etal-2023-synthetic, kurakin2023harnessing, mattern-etal-2022-differentially, putta2023differentially}.
Differential privacy (DP) offers an appealing solution, as it provides a theoretical guarantee of privacy preservation that is controllable through a specified privacy budget.
Although the bulk of work in developing DP approaches has been centered around models trained on tabular and image-related data, there has been increasing interest in applying DP to unstructured text data \cite{shi-etal-2022-selective, yue-etal-2021-differential, DBLP:journals/corr/abs-1910-08902}.

While initial results of synthetic data are promising, prior work has lacked grounding in realistic applications, for example, running experiments with public internet data that language models may already have been exposed to during pre-training. 

In contrast, we conduct experiments on text data from two high stakes domains: healthcare and child protective services, and we rigorously evaluate the synthesized text for its utility, privacy, and potential fairness implications.
For utility and privacy, we introduce novel well-motivated evaluation criteria (``silver'' coreference modeling and entity-centric metrics).
To the best of our knowledge, no prior work has investigated fairness considerations in this domain.

We evaluate several approaches for privacy-preserving synthetic data generation, including fine-tuned models and an in-context-learning approach.
While these evaluations reveal some promising opportunities for synthetic text, they further expose utility degradation, privacy leakage (even when using DP), and issues with group fairness. These results indicate that prior simplistic evaluations have overestimated current viability of synthetic data.

Our primary contributions include a rigorous and reproducible evaluation framework that exposes limitations underestimated in prior work, and empirical results over real high-stakes data.
Overall, our work demonstrates that contrived metrics do not necessarily translate to more realistic scenarios, emphasizing the need for thorough in-domain evaluation to understand potential strengths and limitations of synthetic data.
\section{Methodology}

\subsection{Text Generation}
The primary goal of privacy-preserving synthetic text generation is to generate realistic, but entirely synthesized text for a high stakes domain, such as doctors' notes from a healthcare institution.
We assume we have a data set of real text from that domain, which we can use to guide the generation.
In addition to being realistic, it needs to be ensured that the synthetic data does not reveal uniquely identifiable information about any individuals from the original data.

\paragraph{Fine-tuning}

We adapt the dominant approach from prior work \citep{yue-etal-2023-synthetic}: starting with a pre-trained autoregressive language model, fine-tune it using the real in-domain data, and then generate new data from it.
We compare fine-tuning the model with and without DP, where we use DP-SGD for differentially private fine-tuning. For reference, we provide background on DP and DP-SGD in \Cref{app:dp_background}.
After fine-tuning, we utilize top-k sampling \cite{fan-etal-2018-hierarchical} and nucleus sampling \cite{nucleus} to generate diverse synthetic notes.

We condition the text generation on \textit{control codes} \citep{keskar2019ctrl}. During training, we prepend one or more labels associated with the text to the model input. We similarly prepend control codes during inference, where we sample the provided codes from their distribution in the training data. Thus, during training and inference, the probability distribution of the subsequent text $x = \{x_{1}, x_{2} ... x_{n}\}$ is conditioned on the control code information $c$, which is described by the following equation:

\begin{equation}
P(x|c) = \prod_{i=1}^{n} P(x_i |x_{1} ... x_{i-1}, c)
\end{equation}

Controllable generation approaches enable the generation of notes with specific properties. We primarily use them to enable classification-based utility evaluations (described in \Sref{sec:utilty_eval}).

\paragraph{ICL} In order to explore the potential capabilities of much larger models and investigate if fine-tuning is actually needed, we also generate notes using in-context learning (ICL).
We provide as context examples of training data text with prepended control codes, followed by an additional set of codes to prompt the model to generate content in accordance with the final set of codes. The number of examples provided varies, as we require that each control code for the note to be generated is associated with at least one in-context example. This approach is most directly comparable to the fine-tuned models without DP.

\subsection{Utility Evaluation}
\label{sec:utilty_eval}
Given the goal of developing synthetic data that could be shared externally with researchers or third-party contractors, we evaluate the data's utility based on the performance of NLP models trained over this data. More specifically, we train NLP models on the synthetic data and evaluate their performance over real data. 

\paragraph{Classification}
Similar to prior work \citep{yue-etal-2023-synthetic,kurakin2023harnessing}, we evaluate model performance over classification tasks, where we use the control codes provided during text generation as class labels. We focus on multiclass and/or multilabel classification tasks, and we compare model performance as task difficulty increases.

\paragraph{Coreference Resolution}
Classification tasks can be highly dependent on keywords and phrases, and they do not necessarily require training data to be coherent and consistent across a full paragraph or document. Consistency of entity properties across a document, however, is a necessary condition for coreference training data. Coreference and the related task of mention detection also offer a realistic use case in processing expert-written notes \citep{gandhi-etal-2023-annotating}. Thus, we measure the utility of the synthetic data for training coreference models. 

Unlike classification labels, coreference annotations cannot be easily generated through control codes.
In a practical setting, annotations of coreference clusters would likely be conducted over synthesized data manually by hired annotators or researchers, but this process does not scale for evaluating of multiple iterations of synthetic data generators. Instead, we use a fine-tuned coreference model to simulate ``silver'' annotations over the synthesized data.

More specifically, given a subset of the original dataset $D$ annotated with gold coreference clusters, we first fine-tune a pretrained coreference model \citep{kirstain-etal-2021-coreference} on this data. Using this model, we infer coreference clusters over synthetic data from the same domain which we consider silver annotations. We fine-tune a separate coreference model that has not been task-finetuned with the silver coreference clusters to approximate the utility of the synthetic data for coreference resolution.

We run all experiments with a neural coreference model \citep{kirstain-etal-2021-coreference}. We report results after fine-tuning the model for 40 epochs, where scores are averaged over standard coreference metrics: $\text{MUC}, \text{B}^3, \text{CEAF}_{\phi_4}$.

\subsection{Privacy Evaluation}

\paragraph{Canary Attacks}

Consistent with prior work, to assess the potential leakage of sensitive information in our training data and the extent to which the model memorizes personally identifiable information (PII), we use the canary evaluation method proposed by \cite{DBLP:journals/corr/abs-1802-08232}. This approach involves injecting artificial canary sequences containing PII into the training data and analyzing the likelihood of the frequency of appearance of this PII in the generated outputs. 

We create artificial canary samples that are contextually relevant to both domains and include PII such as names, emails, addresses, and numeric identifiers (details in the appendix in \Cref{tab:canary-type-cps} and \Cref{tab:canary-mimic-types}). Following the methodology outlined in \cite{yue-etal-2023-synthetic}, we vary the number of injections of these canary samples into our training data for 1, 10, and 100 repetitions. For each canary, we generate 10,000 candidate sequences and rank the canaries based on their perplexity score.

\paragraph{Entity-focused metrics}

As canary evaluations are only a proxy for assessing potential privacy risks and may not be comprehensive, we directly leverage entity markers in our datasets to evaluate privacy concerns (we provide details on data-specific entity definitions in \Sref{sec:experimental_setup}).

We compare the frequency of identified entities in the original vs.~synthetic data. Further, while an isolated entity poses some privacy risk, the risk is magnified if the context surrounding the entity is also leaked. Thus we examine the frequency of entities with variable-length surrounding context in the synthetic data and compare them with the training data to estimate the number of memorized patterns that reappear in the synthetic data.

\subsection{Fairness Evaluation}

We compute fairness metrics over the same control-code classification tasks as the utility evaluation (\Sref{sec:utilty_eval}). In data with available demographic information, we compare fairness classification for race and gender subgroups using equality difference (ED) and equalized odds (EO) metrics. For ED, for instance, False Positive Equality Difference (FPED) is the sum of the differences between the overall false positive rate (FPR) for the entire dataset and the FPR for each subgroup. EO constitutes a stricter notion of fairness by evaluating whether both the FPR and TPR rates are the same across all groups. In both cases, values closer to zero indicate that the model performs more uniformly across subgroups, with zero indicating perfect parity across subgroups. For reference, we formally define these metrics in \Cref{fair-ref}.

\setlength{\tabcolsep}{5pt}
\renewcommand{\arraystretch}{1.2}

\section{Experimental Setup}
\label{sec:experimental_setup}

\subsection{Data}
\paragraph{Healthcare} Our primary source of healthcare data is the MIMIC-III Clinical Database \citep{johnson2016mimic,johnson2016mimic2,goldberger2000physiobank}, which contains $>2$M  deidentified notes associated with $>40$K patients admitted to the Beth Israel Deaconess Medical Center in Boston, Massachusetts.  

\begin{table*}[ht]
\centering
\begin{tabular}{ccccc}
\hline
\textbf{\begin{tabular}[c]{@{}c@{}}Training\\ Data\end{tabular}} & \textbf{Dataset} & \textbf{\begin{tabular}[c]{@{}c@{}}F1 \\ Micro\end{tabular}} & \textbf{\begin{tabular}[c]{@{}c@{}}F1 \\ Macro\end{tabular}} & \textbf{\begin{tabular}[c]{@{}c@{}}Subset \\ Accuracy\end{tabular}} \\ \hline
\hline
$D_{real}$                       & $\text{ICD-9}_{n=3}$     & 0.89 ± 0.0 & 0.90 ± 0.0 & 0.76 ± 0.002 \\
$D_{\epsilon = \infty}$          & $\text{ICD-9}_{n=3}$     & 0.87 ± 0.003 \da{-0.02} & 0.87 ± 0.005 \da{-0.03} & 0.74 ± 0.004 \da{-0.02} \\
$D_{\epsilon = 8}$               & $\text{ICD-9}_{n=3}$     & 0.85 ± 0.002 \da{-0.04} & 0.85 ± 0.003 \da{-0.05} & 0.71 ± 0.003 \da{-0.05}        \\ \hline
$D_{real}$                       & $\text{ICD-9}_{n=5}$     & 0.77 ± 0.008 & 0.75 ± 0.016 & 0.56 ± 0.007 \\
$D_{\epsilon = \infty}$          & $\text{ICD-9}_{n=5}$     & 0.75 ± 0.003 \da{-0.02} & 0.73 ± 0.003 \da{-0.02} & 0.55 ± 0.004 \da{-0.01} \\
$D_{\epsilon = 8}$               & $\text{ICD-9}_{n=5}$     &  0.68 ± 0.004 \da{-0.09} & 0.60 ± 0.008 \da{-0.15} & 0.48 ± 0.003 \da{-0.08} \\
\hline
$D_{real}$                       & $\text{ICD-9}_{n=10}$    & 0.70 ± 0.010 & 0.67 ± 0.012 & 0.32 ± 0.016 \\
$D_{\epsilon = \infty}$          & $\text{ICD-9}_{n=10}$    & 0.66 ± 0.001 \da{-0.04} & 0.61 ± 0.003 \da{-0.06} & 0.26 ± 0.004 \da{-0.06} \\
$D_{\epsilon = 8}$               & $\text{ICD-9}_{n=10}$    & 0.54 ± 0.007 \da{-0.16} & 0.40 ± 0.004 \da{-0.27} & 0.18 ± 0.005 \da{-0.14} \\
$D_{ICL}$                        & $\text{ICD-9}_{n=10}$    & 0.57 ± 0.011  \da{-0.13} & 0.47 ± 0.014 \da{-0.20} & 0.21 ± 0.008  \da{-0.11} \\ \hline
\end{tabular}
\caption{Difference in performance between models trained on the synthetic data generated with ($D_{\epsilon = 8}$) and without ($D_{\epsilon = \infty}$) DP and the models trained on real data ($D_{real}$) for multilabel ICD code classification with the top 3, 5, and 10 most frequent labels. Performance degradation greatly increases for more complex tasks.}
\label{tab:mimic-classification}
\end{table*}

As control codes we use ICD-9 codes, which are a standardized format for medical conditions that have been human-annotated in MIMIC. Each note can contain multiple possible codes, making our evaluation task multiclass and multilabel. There are $>5000$ unique ICD-9 codes. Thus, we restrict data to notes containing any of the $n$ most frequent ICD-9 codes, where we typically set $n=10$ and report  $n \in {3, 5}$ for some comparisons, similar to \citet{Aziz2021, Huang_2019}. As a result, the fine-tuning data size for the generative models can vary depending on the value of $n$. The dataset splits for the classification tasks are provided in \Cref{data-stats}. To ensure synthetic data is balanced comparably to real data when evaluating fairness, we additionally provide the patient's ethnicity and biological sex as control codes.

For coreference resolution, we use notes from the MIMIC-II Database annotated for coreference as a part of the i2b2/VA Shared-Task and Workshop in 2011 \citep{uzuner2012evaluating}. This data includes 251 train documents, 51 of which we have randomly selected for development, and 173 test documents.

As the MIMIC data is already deidentified, we directly leverage the strings used for deidentification, e.g. \textit{[**Hospital1 18**]}, \textit{[**First Name3 (LF) 2704**]}, in order to conduct entity-centric privacy evaluations. Finally, we note that although the MIMIC-III diagnoses notes are not permissible to be used for training publicly available language models, there remains a possibility that some MIMIC notes may have been indirectly included in the training data through various other sources.

\paragraph{Child Protective Services (CPS)}
We additionally report results over a data set of contact notes from a county-level Department of Human Services (DHS). These notes log contact with families involved in child protective services, and they are written by caseworkers and other service providers. Unlike MIMIC-III, this data set is not deidentified, which makes it a more realistic test data set, but also prevents the data from being publicly accessible. Throughout our work, this data was stored on a secure server with restricted access, in accordance with IRB-approved protocol and a data sharing agreement established with the county.

The full data set contains 3.1M notes, from approximately 2010 to November 23, 2020. As control codes, we use existing metadata, specifically, the ``Contact Source Description'' field, which specifies one of five possible labels for each note (\textit{Case}, \textit{Investigation}, \textit{Transportation Contact}, \textit{Provider} and \textit{Call Screen}).
For coreference resolution, we use a set of 200 notes annotated for coreference by prior work and shared with us by the county \citep{gandhi-etal-2023-annotating}. This data has train/dev/test sets of sizes 100/10/90 notes.
Finally, for entity-centric evaluations, we use a spaCy NER model to identify spans of entities in the text, and we focus on entities likely to contain private identifying information (e.g., names and organizations).

As CPS cases are complex and involve multiple people, the notion of race or gender for a note is less clear than in the MIMIC data. Thus, we do not report fairness results for this data. We also do not report ICL results, as our single secure server did not have sufficient resources for the larger model.

\subsection{Models}

Our primary text generation model is Sheared-LLaMA-1.3B \citep{xia2024sheared}.\footnote{https://huggingface.co/princeton-nlp/Sheared-LLaMA-1.3B} We fine-tune using Low-Rank Adaption (LoRA) \cite{DBLP:journals/corr/abs-2106-09685}, and we use Opacus \cite{yousefpour2022opacus} for DP fine-tuning. We generally set a privacy budget of $\epsilon = 8$, and $\delta=$  1e-5 (considering our relatively small dataset size), and we report some results with $\epsilon = 4$ for comparison. For ICL, we used the instruction-tuned BioMistral-7B\footnote{https://huggingface.co/BioMistral/BioMistral-7B} model.
As the inference for the BioMistral 7B model is compute-intensive, we report results over experiments conducted only on the $\text{ICD-9}_{n=10}$ subset of the MIMIC-III healthcare dataset, and we generate a smaller number of notes (0.6 as many) as compared to data generated from the smaller fine-tuned models.
We have specified the hyperarameters for each of the models used, dataset distributions and additional detail regarding the experimental setup in \Cref{hyper-ref} and \Cref{data-stats}.

\begin{table}[h]
\centering
\begin{tabular}{ccc}
\hline
 & \multicolumn{1}{c}{\textbf{F1 Score}} & \multicolumn{1}{c}{\textbf{Accuracy}} \\ \hline
 \hline
$D_{real}$             & 0.86 ± 0.003 & 0.86 ± 0.003 \\
$D_{\epsilon = \infty}$      & 0.80 ± 0.006 \da{-0.06} & 0.80 ± 0.006 \da{-0.06} \\
$D_{\epsilon = 8}$     & 0.69 ± 0.002 \da{-0.17} & 0.68 ± 0.002 \da{-0.18} \\ 
$D_{\epsilon = 4}$     & 0.65 ± 0.002 \da{-0.21} & 0.65 ± 0.002 \da{-0.21}                    \\ \hline
\end{tabular}
\caption{Difference in performance between models trained on data generated with differential privacy and models trained on real data, evaluated over CPS classification, for varying privacy budgets.}
\label{tab:cps-classification}
\end{table}

\section{Results}

\begin{table*}[]	
\centering			
\begin{tabular}{lcccc}
\hline				
\multirow{2}{*}{\textbf{Training Data}}& \multicolumn{2}{c}{\textbf{Healthcare}} & \multicolumn{2}{c}{\textbf{CPS}} \\\cline{2-5}			
& \textbf{Mention Detection} & \textbf{Coreference} & \textbf{Mention Detection} & \textbf{Coreference} \\\hline\hline		
$D_{real (gold)}$	&	0.799 ± 0.013	&	0.703 ± 0.011	&	0.877 ± 0.004	&	0.789 ± .005	\\
$D_{real (silver)}$	&	0.659 ± 0.121	&	0.552 ± 0.126	&	0.805 ± 0.007	&	0.642 ± 0.008	\\\hline
$D_{\epsilon = \infty}$	&	0.615 ± 0.051	&	0.443 ± 0.062	&	0.753 ± 0.030	&	0.570 ± 0.035 \\

$D_{\epsilon = 8}$	&	0.599 ± 0.043 &	0.438 ± 0.025	&	0.776 ± 0.007	&	0.589 ± 0.009	\\
$D_{ICL}$	&	0.712 ± 0.010	&	0.588 ± 0.022	&	-	&	-	\\\hline
\end{tabular}		
\caption{F1 scores for coreference and mention detection over entities from human-annotated test splits of the CPS and i2b2/VA datasets. All synthetic datasets are annotated with silver labels. There is general performance degradation for synthetic data generated with ($\epsilon = 8$) and without DP ($\epsilon = \infty$) as compared to real data. The performance degradation is more noticeable in coreference metrics than mention detection metrics. 
}
\label{tab:coref}	
\end{table*}

\subsection{Utility}

\paragraph{Overall Classification} Tables \ref{tab:mimic-classification} and \ref{tab:cps-classification} report results for classification tasks for all models, for the healthcare and CPS data respectively. Unsurprisingly, models trained on data generated from DP fine-tuned models generally under-perform models trained on real data or data generated without DP.  Table \ref{tab:mimic-classification} reports performance for varying task complexity by increasing number of labels $n$ for our multilabel ICD-9 code classification task. For simpler tasks, e.g. $\text{ICD-9}_{n=3}$, there is a much smaller performance degradation and the $D_{\epsilon = \infty}$ (F1$\approx$ 0.87) and $D_{\epsilon = 8}$ (F1$\approx$  0.85) models are nearly comparable. In contrast, there is much larger performance degration for the more difficult $\text{ICD-9}_{n=10}$ task, where F1$\approx$  0.61 for $D_{\epsilon = \infty}$  and F1$\approx$ 0.40 for $D_{\epsilon = 8}$.

In the classification task with the CPS data (\Cref{tab:cps-classification}), however, we notice a significant drop in performance for models trained over data generated with DP 
 for both more generous ($D_{\epsilon = 8}$)  and more restricted ($D_{\epsilon = 4}$) privacy budgets. From examining the data, this task is generally more difficult and the associations between the administrative label and the text in the real data can be quite subtle. It is likely that the generative model often fails to pick up on these associations, and noise introduced by DP further masks these subtleties.

\begin{table}[h]
    \centering
    \begin{tabular}{lcccccccccc}
\hline
& & \textbf{Rank} & \textbf{Perplexity} & \\
\hline
\hline
\multirow{4}{*}{\rotatebox[origin=c]{90}{\textbf{Healthcare}}} & Name & 5986 / 3378 & 49.72 / 54.10 \\
& Address & 2276 / 4075 & 43.59 / 62.66 \\
& Number & 902 / 841 & 9.43 / 14.61 \\
& Email & 711 / 1452 & 37.81 / 72.08 \\
\hline
\multirow{4}{*}{\rotatebox[origin=c]{90}{\textbf{CPS}}}  & Name & 3168 / 2306 & 10.62 / 10.33 \\
& Address & 9618 / 9523 & 21.63 / 27.23 \\
& Number & 474 / 1347 & 16.81 / 23.24 \\
& Email & 387 / 5838 & 49.91 / 81.61 \\
    \end{tabular}
    \caption{Rank and perplexity metrics for 10-insertion canary attacks over MIMIC and CPS data (0, 1 and 100 insertions, reported in \Cref{app:privacy_metrics}, are similar). Each column is formatted as $\epsilon = \infty / \epsilon = 8$ . Higher rank and lower perplexity typically indicate decreased risk of leakage. DP reduces but does not eliminate privacy risks for all canaries, and metrics are generally unstable.}
    \label{tab:canary_evaluations}
\end{table}
\paragraph{Overall Coreference}

Table \ref{tab:coref} reports coreference results.
For comparison, we report $D_{real (gold)}$, model performance when trained over gold in-domain data, which represents the best possible performance we can obtain with human annotations and $D_{real (silver)}$, model performance when trained over silver annotated real data. The ~15 point performance difference in F1 between these two setups represents the performance degradation we should expect to see as a result of inevitable cascading errors from the silver annotations.

There notable performance degradation in synthetic data generated both with and without DP as compared to real data, which is much more noticeable in coreference metrics than mention detection metrics. For example, for the healthcare data from $D_{real (silver)}$ to $D_{\epsilon = 8}$ mention detection F1 declines by 0.044 (0.659 to 0.615), whereas coreference F1 declines by 0.109 (0.552 to 0.443). Models trained on data generated with and without DP perform similarly, likely performance decline is dominated by general quality of synthetic data more so than the application of privacy preservation. While both data sets show similar trends, the performance degradation between real and synthetic data is generally worse for MIMIC-III than CPS.

Data generated by the ICL model resulted in higher-performing coreference systems than data generated by the fine-tuned models. It is likely that the larger model outputted generally more coherent data, through the lack of fine-tuning reduces controllability of generation, e.g., as evidence by the lower performance of the ICL model in \Cref{tab:mimic-classification}.

\subsection{Privacy}

\begin{figure}[h]
\begin{center} 
{
\includegraphics[width=\linewidth]{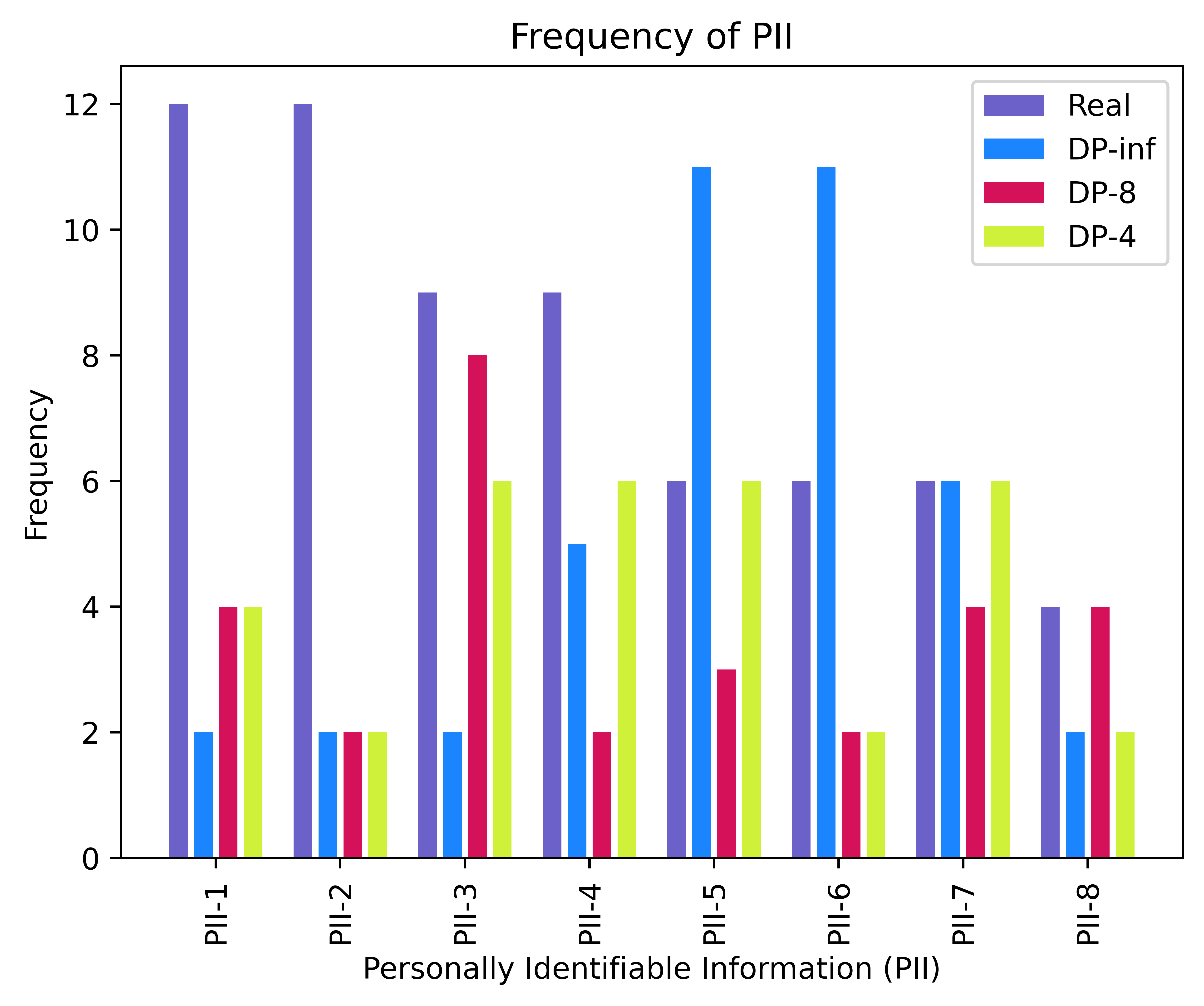}%
\caption{Frequency of the same name (PII-1...PII-8) in real and synthetic CPS data for 8 hand-identified names. The frequency of each name generally, but not always, decreases in synthetic data generated with differential privacy.}
\label{fig:pii-frequency}
}
\end{center}
\end{figure}

\paragraph{Canary Attacks}

\Cref{tab:canary_evaluations} reports results for canary attacks. The DP fine-tuned models exhibit higher perplexity scores for all the canaries, demonstrating that models trained with DP are less likely to output phrases from training data. There is also a relatively sharper drop in perplexity for models fine-tuned without DP as the number of canary insertions increase (\Cref{app:privacy_metrics}). However, our entity-centric evaluation demonstrates that canary evaluations may not be effective in assessing a model's potential for privacy leakage. It should also be noted that while DP improves (increases) rank for some canaries, it decreases rank for others. The canary's rank is also highly dependent on the choice of candidate comparisons, making these metrics easy to skew.

\begin{figure*}
  \begin{minipage}[t]{.5\linewidth}
    \includegraphics[width=\linewidth]{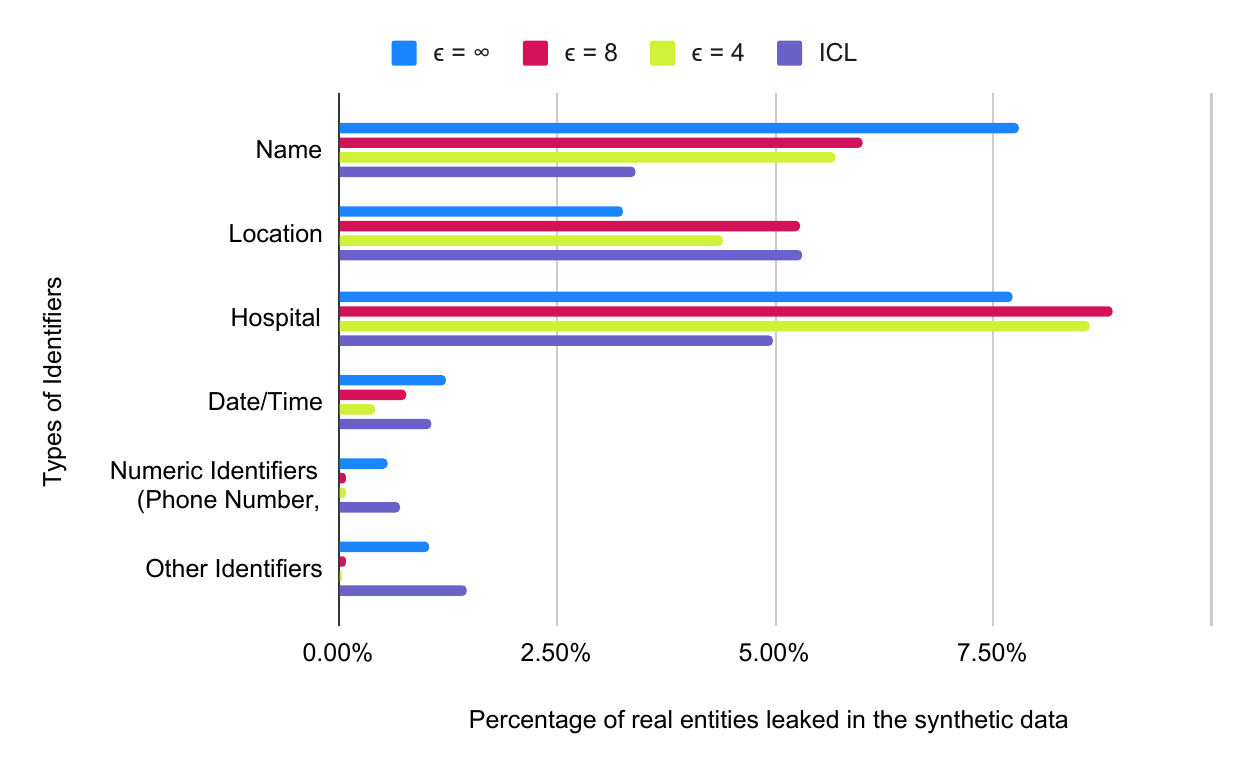}%
  \end{minipage}\hfil
  \begin{minipage}[t]{.5\linewidth}
    \includegraphics[width=\linewidth]{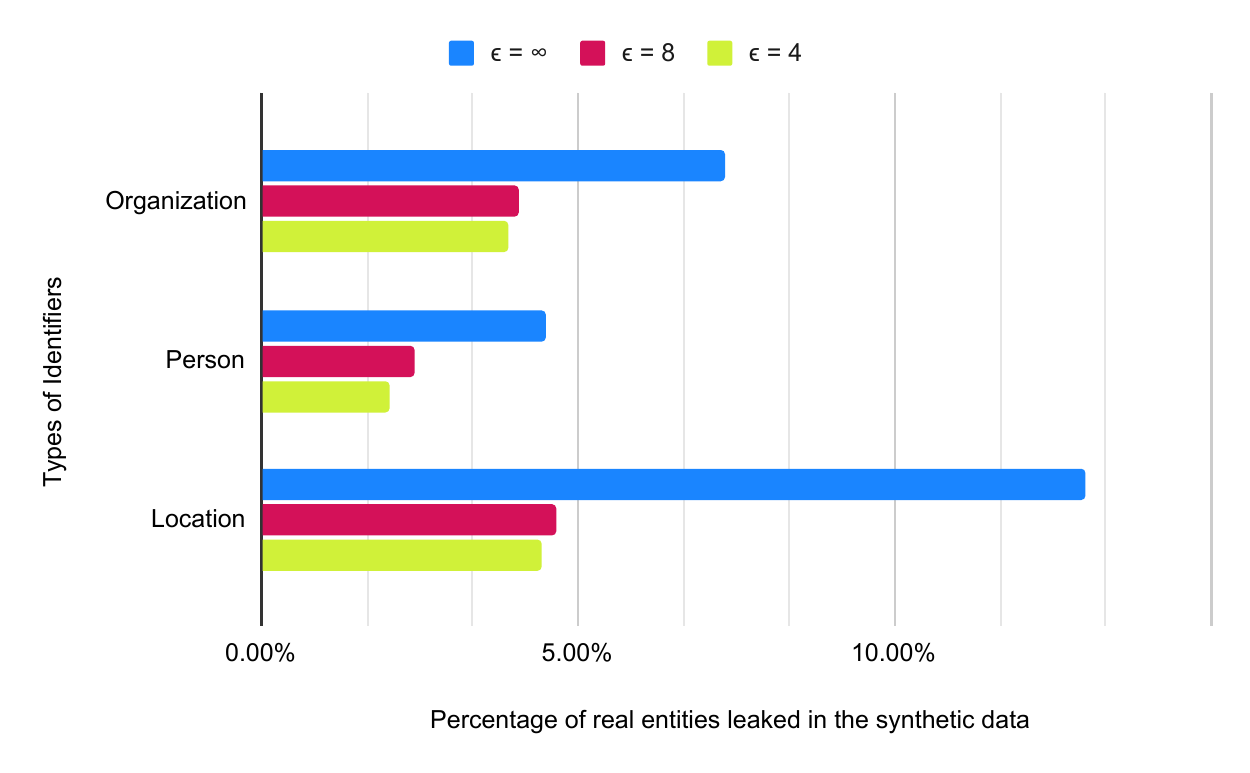}%
  \end{minipage}\hfil
  \caption{Entity-centric privacy evaluation for MIMIC-III (left) and CPS (right). We report the percent of entities in the real data that are present in the synthetic data. While DP reduces leakage, it does not eliminate it entirely for all entities, even with a more restrictive privacy budget.}
  \label{fig:entity_privacy}
\end{figure*}

We perform analysis of actual leakage (e.g., appearance in generated text) using  PII already present in the training data rather than inserted canaries.
These entity-centric metrics (\Cref{fig:entity_privacy}) show that while DP-generated data does contain fewer instances of potentially sensitive information, these entities are not removed from the data entirely, and there is still risk of leakage.

\begin{figure*}
  \begin{minipage}[t]{.33\linewidth}
    \includegraphics[width=\linewidth]{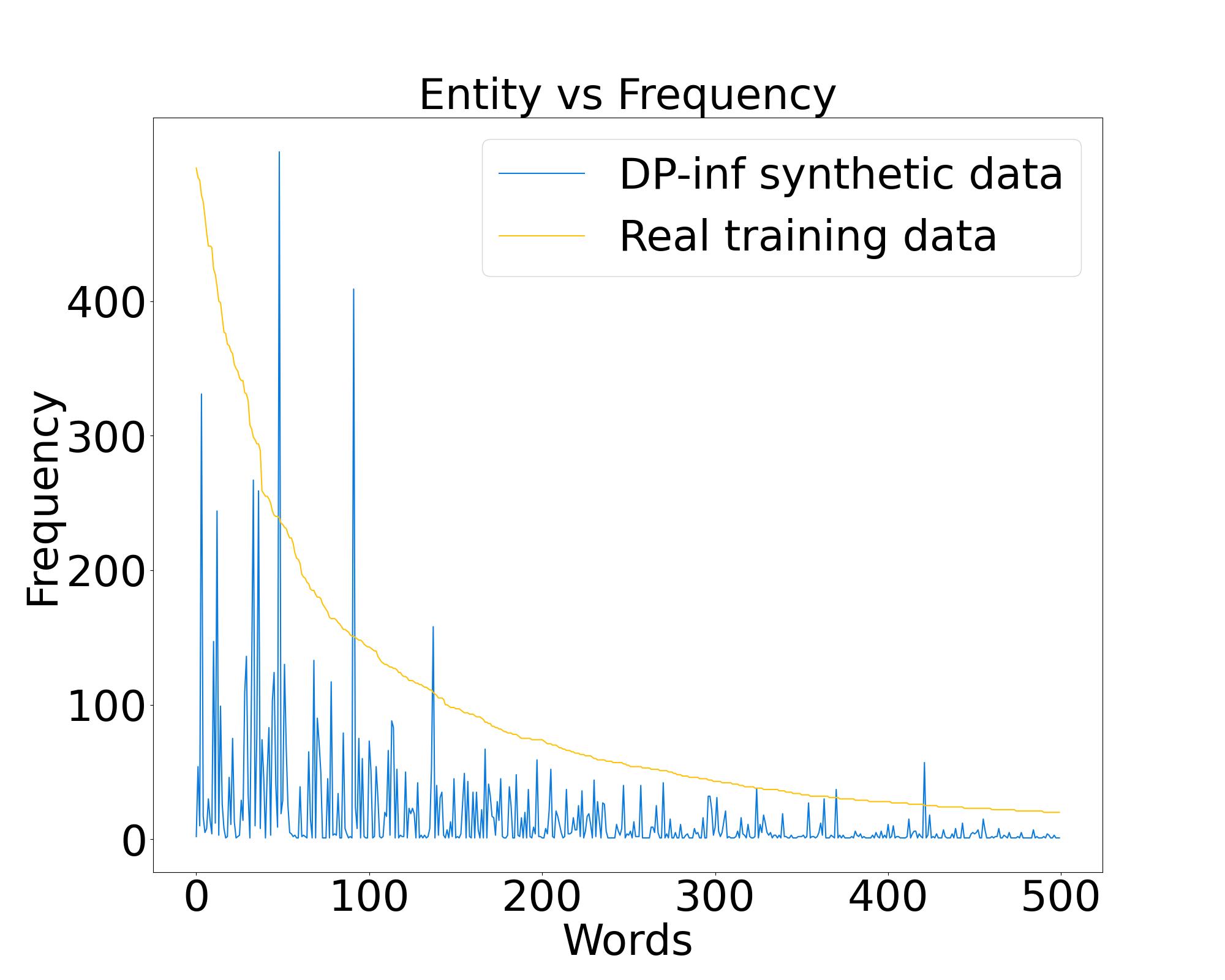}%
  \end{minipage}\hfil
  \begin{minipage}[t]{.33\linewidth}
    \includegraphics[width=\linewidth]{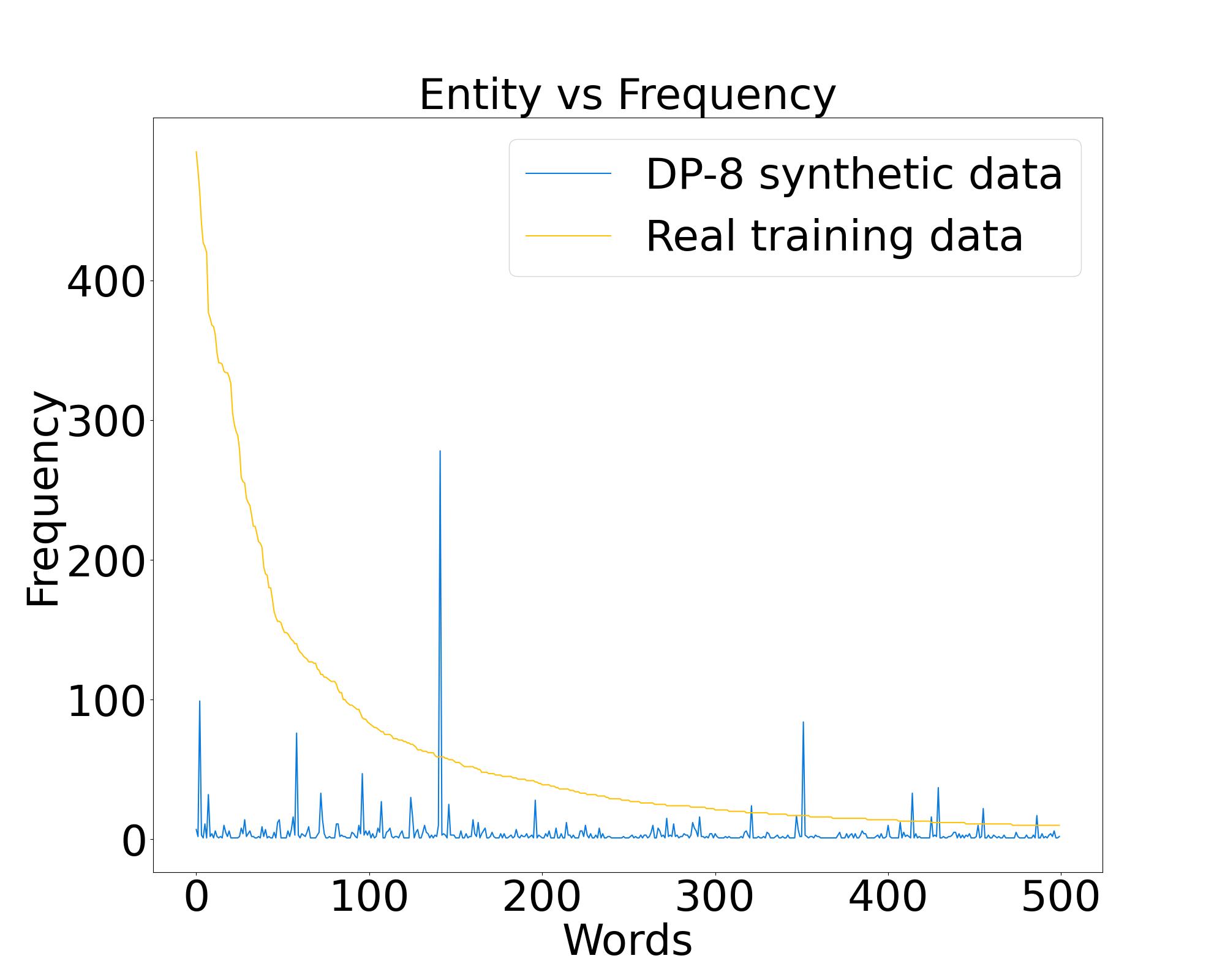}%
  \end{minipage}\hfil
  \begin{minipage}[t]{.33\linewidth}
    \includegraphics[width=\linewidth]{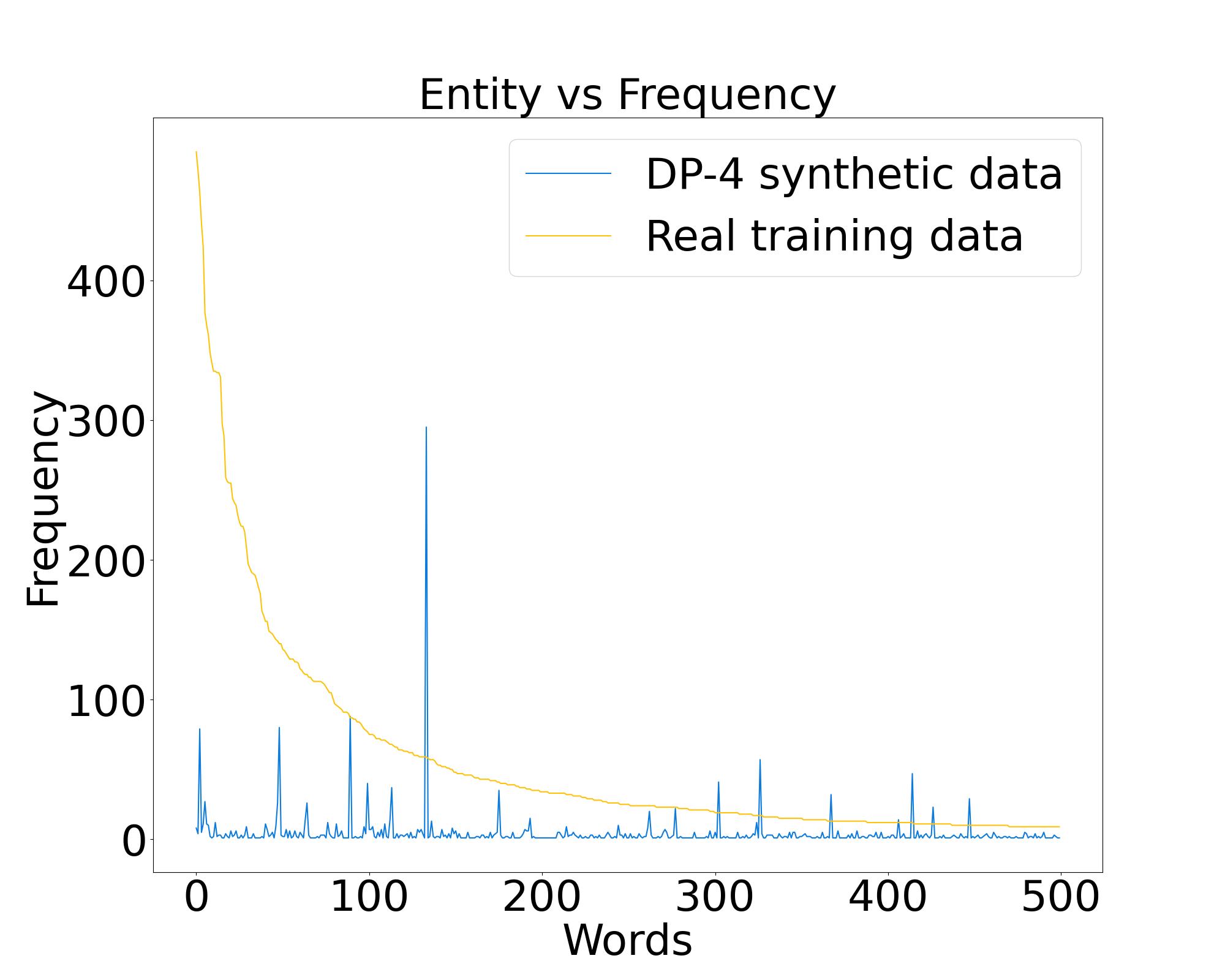}%
  \end{minipage}\hfil
  \caption{MIMIC-III $\text{ICD-9}_{n=10}$ data: Graph depicts the frequency of overlapping entities between the training data $D_{train}$ for the generative model and synthetic data. The top row presents the top 500 most frequent entities from each dataset, limited to entities with a frequency count below 500 in $D_{train}$.}
\label{fig: pvt_entity_freq_mimic}
\end{figure*}

\Cref{fig: pvt_entity_freq_mimic} shows the reduction in private entity leakage in data generated from DP fine-tuned models compared to non-DP fine-tuned models. While notably reduced, some leakage still persists when using DP, even on decreasing the privacy budget further. While \Cref{fig:entity_privacy} compares rates of leakage of aggregated across all entities, it does not provide insight into how leakage for individual entities may occur. In \Cref{fig:pii-frequency} we conduct this analysis by manually selecting 8 names that occur in the real and synthetic CPS data and plotting the frequency of each name in each data type. For most names, frequency in the DP-generated synthetic data is less than in the original data, but this reduction does not always hold, even with a lower privacy budget (e.g., PII-7, PII-5 for $\epsilon = 4$). In data generated without differential privacy, the frequency of names sometimes exceeds their frequency in the original data (e.g., PII-5, PII-6).

Leaked identifiers are potentially more harmful if additional information about an individual is leaked alongside their identity. We assess this risk in \Cref{tab:entity_phrase}, where we gauge how often \textit{sequences} of length 1-4 containing these leaked entities appear in the generated outputs, rather than examining entities in isolation.

The results provide further evidence that, while training models with differential privacy may decrease the risk of information memorization, it does not provide a failsafe. There is a notable disparity in the frequency of phrases from the training data reproduced in these datasets: $D_{\epsilon = \infty}$ contains nearly 1.6 times as many phrases as the $D_{\epsilon = 8}$, but the phrase leakage from $D_{\epsilon = 8}$ is still non-zero. On the other hand, while $D_{ICL}$ is 0.6 times the size of the other datasets, it seems to regurgitate contextual information about these entities from the in-context samples less frequently. However, results from \Cref{fig:entity_privacy} indicate that it still poses privacy risks, as the ICL tends to reproduce these entities, even if not the contexts in which they appear.

\begin{table}[h]
    \centering
    \begin{tabular}{lcccc}

\hline
& \multicolumn{1}{c}{Healthcare} &  & \multicolumn{1}{c}{CPS} \\
      \cline{1-2} \cline{3-4}
 & \textbf{Count} & & \textbf{Count} \\
 \hline
 \hline
$D_{\epsilon = \infty}$    &  16271 & &  4970 \\
$D_{ICL}$                 &  3761  & &  - \\ 
$D_{\epsilon = 8}$        &  10312  & &  1555 \\
$D_{\epsilon = 4}$        & 8934 & &  1434 \\ 
    \end{tabular}
    \caption{Unique contexts in which entities in the real data appear in the synthetic data. Surrounding context word lengths vary from 1 to 4.}
    \label{tab:entity_phrase}
\end{table}

\subsection{Fairness}

We report the FNED and Equalized Odds (EO) metrics for the results from the $\text{ICD-9}_{n = 10}$ multilabel classification tasks in \Cref{tab:mimic_fairness}. The metrics reflect the difference in model performance for the gender and race/ethnicity subgroups with more than 100 samples in the test set, with a larger value indicating more disparate performance across the subgroups. While the gender metrics indicate minimal performance differences, the race/ethnicity metrics show significant disparities. The disparate performance increases for models trained over the data generated from the DP model ($D_{\epsilon = 8}$) as compared to the model without DP ($D_{\epsilon = \infty}$). The model trained with DP ($D_{\epsilon = 8}$) exhibits the most disparate performance across these subgroups, followed by the $D_{ICL}$, although the latter provides better utility for the classification and coreference tasks.

\begin{table}[h]
\centering
\begin{tabular}{cccc}
\hline
                        & & \textbf{FNED}         & \textbf{Equalized Odds} \\ \hline
                        \hline
\multirow{5}{*}{\rotatebox[origin=c]{90}{\textbf{Race}}} &  $D_{real}$       & 0.35 ± 0.0 & 0.20 ± 0.0            \\
& $D_{\epsilon = \infty}$ & 0.39 ± 0.005 & 0.23 ± 0.001  \\
& $D_{\epsilon = 8}$               & 0.53 ± 0.014 & 0.30 ± 0.003  \\
& $D_{ICL}$      & 0.49 ± 0.03 & 0.29 ± 0.012    \\
\hline
\multirow{4 }{*}{\rotatebox[origin=c]{90}{\textbf{Gender}}}  & $D_{real}$       & 0.04 ± 0.0 & 0.04 ± 0.0            \\
& $D_{\epsilon = \infty}$ & 0.02 ± 0.007 & 0.02 ± 0.007           \\
& $D_{\epsilon = 8}$               & 0.04 ± 0.005 & 0.04 ± 0.005           \\
& $D_{ICL}$      & 0.04 ± 0.006 & 0.04 ± 0.006            \\
\hline
\end{tabular}
\caption{Fairness evaluation for the MIMIC-III $\text{ICD-9}_{n=10}$ task, for the gender and race categories. Higher values indicate poorer group fairness performance. We report additional fairness metrics in \Cref{fair-ref} in \Cref{tab:mimic_fairness_extended} that show similar trends.}
\label{tab:mimic_fairness}
\end{table}

\section{Discussion}

Overall, our results are consistent with prior work in that we find only small performance degradation when training a model on DP-generated synthetic text as compared to real data for relatively less fine-grained (e.g. $\text{ICD-9}_{n=3}$, in \Cref{tab:mimic-classification}) classification tasks. Similarly, we do find evidence that DP reduces potential privacy leakage in that artificial canaries (\Cref{tab:canary_evaluations}) and real entities (\Cref{fig:entity_privacy}) are generated less frequently by DP-fine-tuned models.

However, our evaluations also expose previously unexplored weaknesses to this approach. Model performance degrades much more sharply as task complexity increases (e.g. $\text{ICD-9}_{n=10}$ classification in \Cref{tab:mimic-classification}, mention vs.~coreference performance in \Cref{tab:coref}). These results suggest that DP-generated synthetic data may be of sufficient quality for certain NLP tasks and domains, but the quality degradation from DP is a limitation on broader use.

Post-hoc data filtering and re-ranking may offer a way to improve quality. For example, NLI-based approaches have previously been used to rank or evaluate the quality of the generated text \citep{dusek-kasner-2020-evaluating, garneau-lamontagne-2021-trainable, chen-eger-2023-menli} and have been incorporated into the generation pipeline to enhance the consistency of outputs produced by LMs \citep{mersinias-mahowald-2023-generated}, though our initial results with this approach were inconsistent accross data sets.

Furthermore, despite claims that  differentially private training of language models can effectively \textit{eliminate} the risk of privacy leakage \citep{yue-etal-2023-synthetic, mattern-etal-2022-differentially}, our experiments
indicate that there is also a substantial risk of data leakage (Tables \ref{tab:canary_evaluations}-\ref{tab:entity_phrase}, \Cref{fig:entity_privacy}), especially for some types of PII. These results are consistent with risks of leakage identified in sentence-level applications of differential privacy \citep{10179300}.

On investigating the privacy leakage further, we identify several possible causes.
Even for sensitive spans that appear infrequently in the training data, their sub-tokens can recur throughout the same document and across multiple documents more frequently. For instance, in the MIMIC dataset, a token like \textit{[**Hospital1 18**]} might have its "hospital" component repeat multiple times in the data, while the numerical identifier may appear frequently in other contexts, allowing the model to learn all components of the full sensitive span, despite DP-fine-tuning.
This pattern can similarly occur for real identifiers, such as when individuals share the same first name or last name in the CPS data.
Additionally, the presence of sensitive tokens in the pretraining data and the contextual dependencies in text generation may contribute to the model's memorization of sequences in the fine-tuning data.
Finally, correctly defining the unit of privacy presents a significant obstacle in text settings~\citep{chua2024mind}. Ensuring privacy at the user-level is naturally greater than that for a single record, potentially requiring additional utility loss or greater computational costs \citep{charles2024fine}.
Combining privacy-preserving techniques may be a more promising approach than relying on DP.

We further find substantial variance not only in the task difficulty, but also across data sets. Co-reference performance degradation from real to synthetic data is markedly worse for MIMIC than CPS (\Cref{tab:coref}). These differences could be due to a number of factors, such as the similarity between each private data set and the model pre-training data. Regardless, these results emphasize the importance of evaluating on in-domain data, as results may not generalize.

\section{Related Work}

The majority of research on enabling shareable sensitive data has focused on text anonymization, replacing or redacting private information like names and addresses from text. While some approaches redact and replace sensitive information using deterministic rule-based systems \cite{automated, yermilov-etal-2023-privacy, ben-cheikh-larbi-etal-2023-clinical, convo1, volodina-etal-2020-towards}, others employ masked language models \cite{yermilov-etal-2023-privacy}. Differentially private mechanisms have also been integrated into text sanitization processes, such as differentially private  perturbation of text embeddings \citep{10.1145/3336191.3371856} or sampling of replacement tokens \cite{yue-etal-2021-differential, chen-etal-2023-customized} building on the principle of Metric-Local DP \cite{8429310}. Although these methods are computationally inexpensive and domain-agnostic, they have weak privacy guarantees and limited capacity to modify text \cite{mattern-etal-2022-limits, 10.1145/3433638, 10.1145/3531146.3534642}.

Recently, datasets comprised entirely of synthetic data have become potentially viable \citep{8621223, YALE2020244}.
Our work differs from similar approaches to synthetic data generation in its focus on actual high stakes data and thorough grounded evaluation \cite{yue-etal-2023-synthetic, kurakin2023harnessing, mattern-etal-2022-differentially, putta2023differentially}. Notably, \citet{Aziz2021} do similarly investigate healthcare data, but they do not evaluate potential privacy leakage, and their utility measures do not adequately capture errors in text fluency and consistency which is crucial for finer-grained applications.

A separate but overlapping line of work has focused on improving privacy in NLP models \& protection against membership inference, rather than in the generated data.
This work has similarly trained NLP models with differential privacy but has evaluated direct performance of these models on downstream tasks \citep{li2021large, wu-etal-2022-adaptive}. Nevertheless, this line of work is not directly comparable to ours, as it focuses on training models directly on private data, while our approach promotes the shareability of data and imposes fewer restrictions on sharing models trained using private synthetic data.

\section{Conclusions}
Although synthetic data generated with differential privacy is an appealing way to improve responsible AI development, off-the-shelf DP does not achieve sufficient privacy, utility, or fairness over real high-stakes data.
These failings suggest numerous opportunities for future work on improving the coherence of synthetic data and the application of privacy preservation to this task.
Our evaluation methods offer a way to foster this research, with grounding in real applications, rather than contrived settings, where performance is liable to being over-estimated.

\section*{Acknowledgments}
This work was support in part by the JHU + Amazon Initiative for Interactive AI (AI2AI). We also thank the anonymous county Department of Human Services for providing feedback and data for this work, and we thank Danish Pruthi, Krishna Pillutla, and Jessica Sorrell for their helpful feedback.

\section{Limitations}

The primary limitation of our work is the impossibility of considering all possible model and parameter configurations. While we selected high-performing models that we were able to fine-tune and evaluate on our compute resources, results may differ for different pre-trained language models. Similarly, while we select hyper-parameters based on prior work and conduct some ablation studies, text-generation is extremely compute-intensive and a fully exhaustive hyper-parameter sweep is not feasible. Overall our results emphasize the need to thoroughly evaluate models on target data and cannot necessarily be assumed to generalize to untested data.

There are also additional approaches we do not explore that could reduce privacy risk or improve the quality of synthetic data generated during training.
Examples include combining text-anonymization with DP fine-tuning or selective constraints applied to the training data to reduce the frequency of entity mentions. However, this is difficult in practice, as real-world data is complex with, for example, the same people mentioned across multiple CPS cases. 

\section{Ethical Considerations}
Our work involves the use of private sensitive data, particularly the CPS data, which is not de-identified. To minimize risk, throughout this project we maintained a high level of data security, in compliance with IRB-approved protocol. The CPS data was exclusively stored on a secure restricted-access server with HIPPA-standard of security. All CPS experiments were conduct on this server, which also limited the models we could investigate. Our paper does not include any examples from either data set, in compliance with their respective data use agreements.

\bibliography{anthology,custom}

\appendix

\section{Background: Differential Privacy}
\label{app:dp_background}

Differential privacy offers a formal privacy guarantee that ensures that any individual's data cannot be inferred from a query applied to a dataset \citep{Dwork2006,dwork2014algorithmic}. In other words, the result of such a query is nearly indistinguishable from the result of the same query applied to a dataset that either includes a modified version of the individual's data or excludes the record entirely, thereby preserving the individual's privacy. In this case, the notion of adjacency is defined as a difference of a single record in the original dataset D and the modified dataset D'. 

Formally, differential privacy is defined as follows:

\textbf{Definition}: Given a dataset \(D\) and an adjacent dataset \(D'\), which is produced by removing or modifying a single record from \(D\), a randomized algorithm \(F : D \to Y\) is \((\epsilon, \delta)\)-private if for any two neighboring datasets \(D, D'\), with the constraints \(\epsilon > 0\) and \(\delta \in [0, 1]\), the following holds true for all sets \(y\) \(\subseteq\) \(Y\):
\[
\Pr[F(D) \in y] \leq e^{\epsilon} \Pr[F(D') \in y] + \delta
\]
The value of \(\epsilon\) denotes the privacy budget, while \(\delta\) specifies the likelihood that the privacy guarantee may fail. If \(\delta\) is set to 0, this implies a purely differentially private setting with no probability of the guarantee being broken. The value of \(\epsilon\) constrains how similar the outputs of both distributions are; a higher \(\epsilon\) value indicates a greater privacy budget, meaning the algorithm is less private. DP guarantees that even if an adversary has access to any side-knowledge, the privacy leakage of \((\epsilon, \delta)\)-DP algorithms will not increase. Additionally, another property of DP is that it ensures that any post-processing on the outputs of \((\epsilon, \delta)\)-differentially private algorithms will remain \((\epsilon, \delta)\)-differentially private. 

We use DP-SGD \citep{Abadi2016DPSGD}, a modification to the stochastic gradient descent (SGD) algorithm, which is typically used to train neural networks. DP-SGD clips the gradients to limit the contribution of individual samples from the training data and subsequently adds noise from a predefined type of distribution (such as a Gaussian or Laplacian distribution) to the sum of the clipped gradients across all samples. DP-SGD thus provides a differentially private guarantee to obfuscate the gradient update, thereby ensuring that the contribution of any given sample in the training data is indistinguishable due to the aforementioned post-processing property. This process ensures \((\epsilon, \delta)\)-differential privacy for each model update. Given a privacy budget, number of epochs, and other training parameters, we can estimate the privacy parameters using estimation algorithms \cite{gopi2021numerical}.

\section{Hyperparameters}\label{hyper-ref}

For training the autoregressive model, we used an effective batch size of 32 for training the non-differentially private model for both the CPS and the MIMIC-III data. For the differentially private fine-tuning, we used an effective batch size of 1024. We set the maximum sequence length to 1024 tokens and and our training was conducted over 3 epochs, and training was optimized using the AdamW optimizer with its default hyperparameters. For the MIMIC-III data, our learning rate was set to 3e-4 (for the non-DP finetuned models) and to 1e-3 (for the DP-finetuned models). For the CPS data, we found a learning rate of 3e-4 in both cases was optimal. For the LoRA hyperparameters, we used a dimension of 4 and an alpha value of 32, specifically targeting the query (q\_proj) and value (v\_proj) projection layers of the transformer. To ensure training stability, we applied gradient clipping with a maximum gradient norm of 1.0. For the DP fine-tuning of the autoregressive model, we train with a privacy budget of epsilon = 8 for most of our experiments, and considering our relatively small dataset size we set delta to 1e-5 for our experiments. 

For training the downstream classifier, we conducted training over 3 epochs with a batch size of 8 and a maximum sequence length of 512 tokens. We utilized the AdamW optimizer with a learning rate of 5e-5. We also conducted these downstream experiments with RoBERTa and found that the differences were minimal, with no impact on the overall trends, so we decided not to include these results.

During inference, we set the top-k sampling parameter to k = 50 and the nucleus sampling parameter to p = 0.95. We generate approximately 30k and 31k samples for the child welfare data and diagnosis notes for the 10 most frequent ICD-9 codes, respectively, which are then used to train the downstream classifiers. We use similar inference hyperparameters for the instruction-tuned BioMistral-7B model for ICL, we set the top-k value to 50, top-p  to 0.9 and the penalty-alpha parameter to 0.6. 

Our experiments for all the aforementioned experimental setups used an A100 GPU for the MIMIC data and A6000 GPUs on a single secure server for the CPS data.

\begin{table*}[h]
\centering
\begin{tabular}{cccccc}
\hline
                        & \textbf{FNED}         & \textbf{FPED}         & \textbf{TPED}         & \textbf{TNED}         & \textbf{Equalized Odds} \\ \hline
\textbf{Race}           &                       &                       &                       &                       &                         \\
$D^{base}_{real}$              & 0.35 ± 0.0 & 0.01 ± 0.0 & 0.35 ± 0.0 & 0.01 ± 0.0 & 0.20 ± 0.0 \\
$D_{\epsilon = \infty}$ & 0.39 ± 0.005 & 0.02 ± 0.003 & 0.39 ± 0.005 & 0.02 ± 0.003 & 0.23 ± 0.001 \\
$D_{\epsilon = 8}$      & 0.53 ± 0.014 & 0.01 ± 0.003 & 0.53 ± 0.014 & 0.01 ± 0.003 & 0.30 ± 0.003          \\
$D_{ICL}$               & 0.49 ± 0.03 & 0.03 ± 0.006 & 0.49 ± 0.03 & 0.03 ± 0.006 & 0.29 ± 0.012 \\ \hline
\textbf{Gender}         &                       &                       &                       &                       &                         \\
$D^{base}_{real}$              & 0.04 ± 0.0 & 0.0 ± 0.0 & 0.04 ± 0.0 & 0.0 ± 0.0 & 0.042 ± 0.0 \\
$D_{\epsilon = \infty}$ & 0.02 ± 0.007 & 0.0 ± 0.0 & 0.02 ± 0.007 & 0.0 ± 0.0 & 0.02 ± 0.007 \\
$D_{\epsilon = 8}$          & 0.04 ± 0.005 & 0.0 ± 0.001 & 0.04 ± 0.005 & 0.0 ± 0.001 & 0.04 ± 0.005 \\
$D_{ICL}$               & 0.04 ± 0.006 & 0.0 ± 0.002 & 0.04 ± 0.006 & 0.0 ± 0.002 & 0.04 ± 0.006 \\ \hline
\end{tabular}
\caption{Fairness evaluation for the MIMIC-III $\text{ICD-9}_{n=10}$ task, for the gender and race categories.}
\label{tab:mimic_fairness_extended}
\end{table*}

\section{Fairness}\label{fair-ref}

The False Positive Equality Difference (FPED) metric is the sum of the differences between the overall false positive rate (FPR) for the entire dataset and the FPR for each subgroup \( d \in D \), where $D$ is a set consisting of all subgroups corresponding to a demographic attribute within the dataset.

\begin{equation}
\text{FPED} = \sum_{d=1}^{D} \left| \text{FPR}_{\text{overall}} - \text{FPR}_{d} \right|
\end{equation}

\begin{equation}
\text{TNED} = \sum_{d=1}^{D} \left| \text{TNR}_{\text{overall}} - \text{TNR}_{d} \right|
\end{equation}

Similarly, these ED metrics can be estimated for the true positive, true negative and false negative rates to estimate the TPED, TNED and FNED respectively. Lower values of these ED scores indicate that the model's performance is more consistent across different subgroups.

The Equalized Odds ratio is calculated as follows: 

\begin{multline*}
\text{EO}_{D} = \max \Biggl( \max_{i \in D}(\text{TPR}_i) - \min_{i \in D}(\text{TPR}_i), \\
\max_{i \in D}(\text{FPR}_i) - \min_{i \in D}(\text{FPR}_i) \Biggr)
\end{multline*}

We have two categories of subgroups that are present in the MIMIC-III dataset over which we perform fairness evaluations with the downstream classifier trained over synthetic data with demographic control codes. The following categorical variables assigned to each within the dataset:

\begin{itemize}
    \item \textbf{Gender:} Female, Male
    \item \textbf{Race/Ethnicity:} American Indian/Alaska Native, Asian, Black, Hispanic/Latino, Middle Eastern, Multi Race/Ethnicity, Other, Portuguese, South American, White
\end{itemize}

The format of the control code for the MIMIC-III data is as follows: \textit{Long\_Title: <diagnoses>, ICD9\_CODE: <codes>, Gender: <gender>, Ethnicity: <ethnicity>}, where the <diagnoses> variable represents the long title form of the ICD-9 codes, information that is already provided with the MIMIC-III dataset.

\section{Data Statistics}\label{data-stats}

Our train/dev splits for the CPS, $\text{ICD-9}_{n=10}$, $\text{ICD-9}_{n=5}$ and $\text{ICD-9}_{n=3}$  datasets the generative model was trained on are 89327/4701, 44215/2327, 37245/1960, 31317/1648 respectively.

The size of the train/dev sets for the models trained for downstream classification on the real ($D_{real}$) and synthetic ($D_{\epsilon = \infty,\ 8,\ 4}$) CPS data is 25385/2821, and the test set for this task consists of 4949 records.

For the $\text{ICD-9}_{n=10}$ multilabelling task, the real ($D_{real}$) and synthetic ($D_{\epsilon = \infty,\ 8,\ 4}$) train/dev split was the same, with $\simeq$ 27920/3100 for all models, and the test set size was $\simeq$ 7500 samples. For the $\text{ICD-9}_{n=5}$ task, the train/dev split was the same for all models $\simeq$ 23520/2615, and the test set size was $\simeq$ 6315 samples. Similarly, for the $\text{ICD-9}_{n=3}$ task, the train/dev split was $\simeq$ 19780 / 2200, and the test set size was $\simeq$ 5310 samples. Each of these experiments for the downstream tasks (coreference/mention detection \& classification) was averaged over 3 runs.

\section{Extended Privacy Evaluation results}
\label{app:privacy_metrics}

In \Cref{tab:canary_evaluations} we report the full set of canary results (for 1, 10, and 100 insertions, for each canary type). Results are generally similar across different numbers of insertions, in that DP generally reduces rank and perplexity, thus improving privacy, but does not eliminate all risk of leakage. 

\begin{table*}[h]
    \centering
    \begin{tabular}{lcccccccccc}
      & \multicolumn{2}{c}{MIMIC} & & \multicolumn{2}{c}{CPS} \\
    \cline{2-3} \cline{5-6}
& \textbf{Rank} & \textbf{Perplexity} & & \textbf{Rank} & \textbf{Perplexity} \\
\hline
\hline

\textbf{100 Insertions} \\
Name & 4628 / 3356 & 35.80 / 53.19 &  & 3025/2265 & 7.08/10.22 \\
Address & 5 / 3967 & 16.52 / 61.37 &  & 57/9401 & 13.01/26.49 \\
Number & 1 / 818 & 5.77 / 14.46 &  & 4/1285 & 9.25/22.88 \\
Email & 1 / 1410 & 10.50 / 70.26 &  & 1/5301 & 7.64/78.42 \\
\hline
\textbf{10 Insertions} \\
Name & 5986 / 3378 & 49.72 / 54.10 &  & 3168 / 2306 & 10.62 / 10.33 \\
Address & 2276 / 4075 & 43.59 / 62.66 &  & 9618 / 9523 & 21.63 / 27.23 \\
Number & 902 / 841 & 9.43 / 14.61 &  & 474 / 1347 & 16.81 / 23.24 \\
Email & 711 / 1452 & 37.81 / 72.08 &  & 387 / 5838 & 49.91 / 81.61 \\
\hline
\textbf{1 Insertion} \\
Name & 6037 / 3383 & 52.06 / 54.20 &  & 3164 / 2320 & 11.27 / 10.34 \\
Address & 3348 / 4081 & 54.50 / 62.79 &  & 9715 / 9529 & 24.65 / 27.29 \\
Number & 1084 / 838 & 9.82 / 14.63 &  & 1016 / 1357 & 20.78 / 23.27 \\
Email & 1941 / 1457 & 43.90 / 72.28 &  & 1771 / 5881 & 62.44 / 81.88 \\ \hline
\textbf{0 Insertions} \\
Name & 6086 / 5265 & 44.81 / 57.54 &  & 3690 / 2321 & 11.89 / 10.34 \\
Address & 4565 / 3869 & 75.79 / 61.68 &  & 9890 / 9529 & 23.15 / 27.29 \\
Number & 1217 / 1522 & 11.80 / 13.33 &  & 1176 / 1357 & 21.27 / 23.28 \\
Email & 1003 / 3174 & 43.80 / 55.19 &  & 5893 / 5881 & 79.05 / 81.88 \\
    \end{tabular}
    \caption{Rank and perplexity metrics for canary attacks over MIMIC and CPS data. Each column is formatted as $\epsilon = \infty / \epsilon = 8$ . Perplexity scores suggest that DP reduces privacy metrics for all canaries, and generally show similar privacy improvements.}
    \label{tab:canary_evaluations_app}
\end{table*}

\begin{table}[h]
\centering
\resizebox{\columnwidth}{!}{
\begin{tabular}{ll}
\hline
\textbf{Type} & \textbf{Canary Sequence}                                                                                                                                                                  \\ \hline
Name          & \begin{tabular}[c]{@{}l@{}}Pt \underline{Robin Kramer} is has been admitted to \\ the hospital and the patient's medical records were \\ updated yesterday.\end{tabular} \\
Address       & \begin{tabular}[c]{@{}l@{}}The patient has been admitted to the hospital at \underline{789} \\ \underline{Elmford St, Maplegreen}.\end{tabular}             \\
Number        & \begin{tabular}[c]{@{}l@{}}The patient's contact number for emergencies is \\ \underline{(541) 545-6789}.\end{tabular}                                                   \\
E-mail        & \begin{tabular}[c]{@{}l@{}}In case of any inquiries, the patient can be contacted at \\ \underline{ptcareastps456@email.com}.\end{tabular}                               \\ \hline
\end{tabular}}
\caption{The canaries inserted into the training data for the models fine-tuned to generate synthetic MIMIC-III data.}
\label{tab:canary-mimic-types}
\end{table}

\begin{table}[h]
\centering
\resizebox{\columnwidth}{!}{
\begin{tabular}{ll}
\hline
\textbf{Type} & \textbf{Canary Sequence}                                                                                                                                            \\ \hline
Name          & \begin{tabular}[c]{@{}l@{}}\underline{Robin Kramer} has been placed in temporary foster \\ care and there is an ongoing investigation into the \\ child's welfare.\end{tabular} \\
Address       & \begin{tabular}[c]{@{}l@{}}The CW visited the foster family's address at \\ \underline{456 W Oak Avenue, Springfield, IL}.\end{tabular}                                         \\
Number        & \begin{tabular}[c]{@{}l@{}}The case number \underline{CW-2023-56893} has been \\ assigned for tracking purposes.\end{tabular}                                                   \\
E-mail        & \begin{tabular}[c]{@{}l@{}}The CW can contact the foster family at \\ \underline{randuser789@xyzreportnews.com} in \\ case of any emergencies.\end{tabular}                     \\ \hline
\end{tabular}}
\caption{The canaries inserted into the training data for the models fine-tuned to generate synthetic CPS data.}
\label{tab:canary-type-cps}
\end{table}

\begin{table*}[ht]
\centering
\resizebox{1.75\columnwidth}{!}{
\begin{tabular}{cccccc}
\hline
\textbf{Model} & \textbf{Data Size} & \textbf{\begin{tabular}[c]{@{}c@{}}Phrase Overlap\\ Ratio\end{tabular}} & \textbf{\begin{tabular}[c]{@{}c@{}}Total \#\\of Phrase Overlap\end{tabular}} & \textbf{\begin{tabular}[c]{@{}c@{}}Total \#\\of Phrases\end{tabular}} & \textbf{\begin{tabular}[c]{@{}c@{}}Total \# of Deidentified \\ Phrases Generated\end{tabular}} \\ \hline
$D_{(real,\ \text{ICD-9}_{\ n=10})}$  & 44215              & 1                                                                       & 2935955                                                                 & 2935955              & 3845112                                                                                  \\
$D_{(\epsilon = \infty,\ \text{ICD-9}_{n=10})}$         & 31020              & 0.00504                                                                & 16271                                                                   & 3229278              & 369390                                                                                   \\
$D_{(\epsilon = 8,\ \text{ICD-9}_{n=10})}$           & 31020              & 0.0314                                                                & 10312                                                                    & 3281866              & 390956                                                                                   \\
$D_{(ICL,\ \text{ICD-9}_{n=10})}$            & 19640              & 0.00117                                                                & 3761                                                                    & 3205098              & 316905                                                                                   \\
               &                    &                                                                         &                                                                         &                      &                                                                                          \\
$D_{(real,\ \text{ICD-9}_{n=5})}$  & 37245              & 1                                                                       & 2565699                                                                 & 2565699              & 3352588                                                                                  \\
$D_{(\epsilon = \infty,\ \text{ICD-9}_{n=5})}$         & 26136              & 0.00478                                                                & 13537                                                                   & 2831658              & 323963                                                                                   \\
$D_{(\epsilon = 8,\ \text{ICD-9}_{n=5})}$           & 26136              & 0.00263                                                                & 7447                                                                    & 2831627              & 295290                                                                                   \\
               &                    &                                                                         &                                                                         &                      &                                                                                            \\
               \hline
\end{tabular}}
\caption{Analysis for the MIMIC-III dataset of all the unique contexts in which entities of from all categories from the training data appear in the synthetic data, considering surrounding context word lengths varying from 1 to 4. $D_{real}$ corresponds to the training data the generative models were trained on.}
\label{tab:context-priv-mimic}
\end{table*}

\begin{table*}[]
\centering
\resizebox{1.75\columnwidth}{!}{
\begin{tabular}{cccccc}
\hline
\textbf{Model} & \textbf{\begin{tabular}[c]{@{}c@{}}Data \\ Size\end{tabular}} & \textbf{\begin{tabular}[c]{@{}c@{}}Phrase Overlap \\ Ratio\end{tabular}} & \textbf{\begin{tabular}[c]{@{}c@{}}Total \#\\ of Phrase Overlap \#\end{tabular}} & \textbf{\begin{tabular}[c]{@{}c@{}}Total \# of Phrases in $D_{real}$  \\ + $D_{synth-data}$ \end{tabular}} & \textbf{\begin{tabular}[c]{@{}c@{}}Total \# of Phrases in \\$D_{synth-data}$ \end{tabular}} \\ \hline
$D_{\epsilon = \infty}$         & 28206                                                         & 0.01517                                                               & 6307                                                                   & 415685                                                                 & 104153                                                                 \\
$D_{\epsilon = 8}$           & 28206                                                         & 0.00619                                                                & 2448                                                                   & 395303                                                                 & 65990                                                                  \\
$D_{\epsilon = 4}$           & 28206                                                         & 0.00567                                                                & 2218                                                                   &    391313                                                              & 59286                                                                 \\ \hline
\end{tabular}}
\caption{Analysis for the CPS data of all the unique contexts in which entities of from all categories from the training data appear in the synthetic data, considering surrounding context word lengths varying from 1 to 4. $D_{real}$ corresponds to the training data the generative models were trained on.}
\label{tab:context-cps-priv-all}
\end{table*}

\end{document}